\documentclass[letterpaper, 10 pt, conference]{ieeeconf}  
\usepackage{mathrsfs}
\usepackage{graphicx}
\usepackage{makecell} 
\usepackage{amsmath}
\usepackage{algorithmic}
\usepackage{algorithm}
\usepackage{stfloats}
\usepackage{float}
\usepackage{makecell}
\usepackage{graphicx}
\usepackage{textcomp}
\usepackage{array} 
\usepackage{longtable}
\usepackage{booktabs}
\usepackage{float}  
\usepackage{multicol}
\usepackage{multirow}
\usepackage{graphicx}
\usepackage{makecell}
\usepackage{amssymb}
\usepackage{stfloats}
\usepackage[table]{xcolor}
\definecolor{lightred}{RGB}{255, 200, 200}
\usepackage{hyperref}
\hypersetup{hypertex=true,
colorlinks=true,
linkcolor=blue,
anchorcolor=blue,
citecolor=blue}

\IEEEoverridecommandlockouts             

\definecolor{myColor}{RGB}{0, 0, 0}

\author{Tianxing Zhou$^{1,2}$, Zhirui Wang$^{1\dag}$, Haojia Ao$^{1\dag}$, Guangyan Chen$^{1}$, Boyang Xing$^{3}$, Jingwen Cheng$^{3}$,\\Yi Yang$^{1}$, Yufeng Yue$^{1*}$
\thanks{$^*$This work was supported by National Key RD Program of China (Grant No.2024YFB4708900).}
\thanks{$^*$ Corresponding author: Yufeng Yue (\tt\small{yueyufeng@bit.edu.cn})}
\thanks{$^\dag$ Equal Contribution}
\thanks{$^{1}$
School of Automation, Beijing Institute of Technology, Beijing, 100081,
China}%
\thanks{$^{2}$
Beijing Zhongguancun Academy, Beijing, 100094, China}%
\thanks{$^{3}$
Humanoid Robot (Shanghai) Co., Ltd., Shanghai, 200093, China}%
}
\title{\LARGE \bf
STEP Planner: Constructing cross-hierarchical subgoal tree as an embodied long-horizon task planner}
\begin{document}

\let\oldtwocolumn\twocolumn

\renewcommand\twocolumn[1][]{%
\begin{figure*}
    \oldtwocolumn[{#1}{
    \begin{center}
        \centering
        \includegraphics[width=0.8\linewidth]{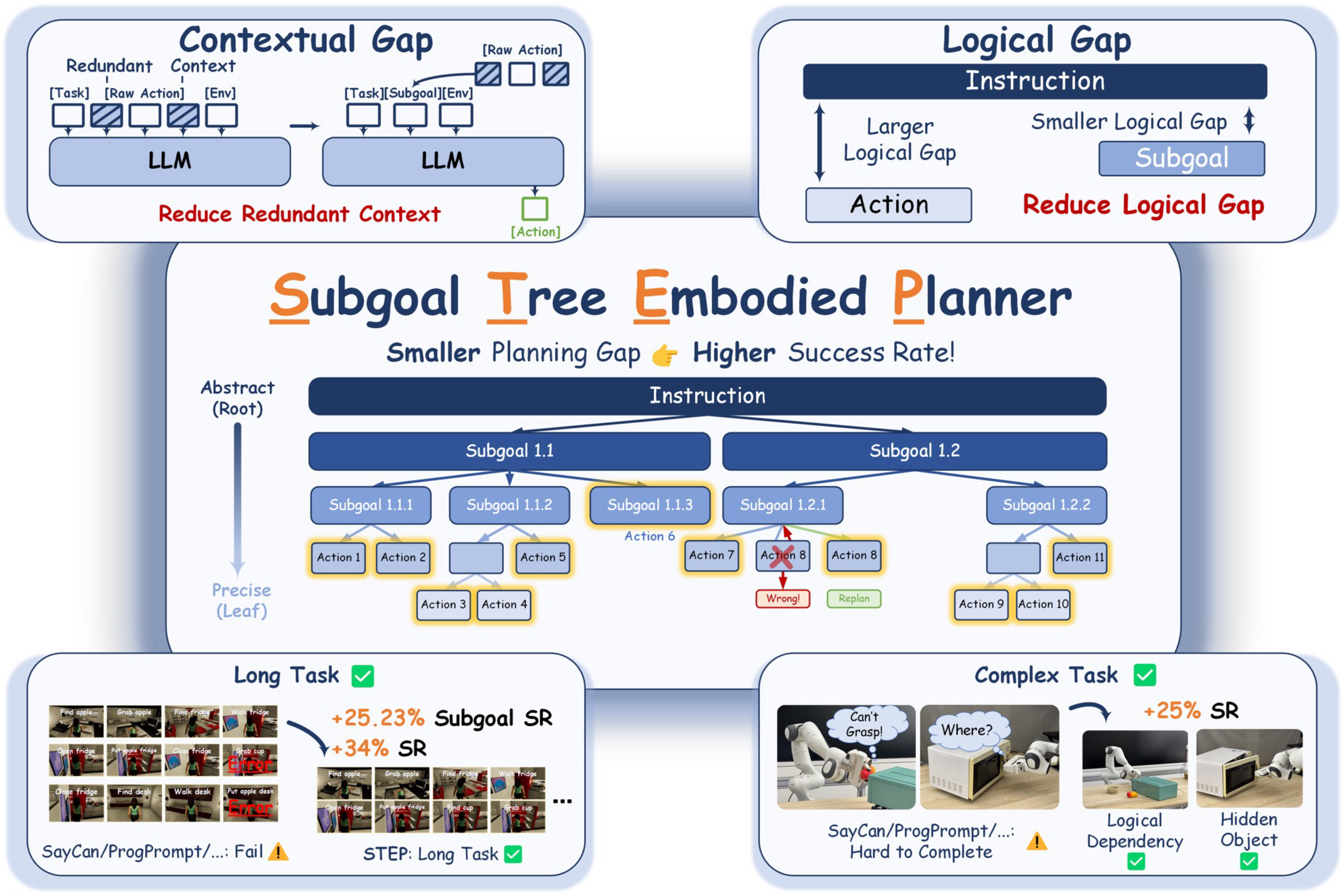}
        \caption{\textbf{Illustration of our STEP.} Through constructing a hierarchical tree structure that decomposes tasks from coarse to fine granularity, STEP effectively bridges both logical and contextual gaps in the planning process, demonstrating superior performance compared to existing methods in long-horizon and complex tasks.
}
        \label{fig:0}
    \end{center}
    }]
\end{figure*}
}

\bibliographystyle{unsrt}

\maketitle


\begin{abstract}

The ability to perform reliable long-horizon task planning is crucial for deploying robots in real-world environments. However, directly employing Large Language Models (LLMs) as action sequence generators often results in low success rates due to their limited reasoning ability for long-horizon embodied tasks.
In the STEP framework, we construct a subgoal tree through a pair of closed-loop models: a subgoal decomposition model and a leaf node termination model. Within this framework, we develop a hierarchical tree structure that spans from coarse to fine resolutions. The subgoal decomposition model leverages a foundation LLM to break down complex goals into manageable subgoals, thereby spanning the subgoal tree. The leaf node termination model provides real-time feedback based on environmental states, determining when to terminate the tree spanning and ensuring each leaf node can be directly converted into a primitive action.
Experiments conducted in both the VirtualHome WAH-NL benchmark and on real robots demonstrate that STEP achieves long-horizon embodied task completion with success rates up to 34\% (WAH-NL) and 25\% (real robot) outperforming SOTA methods.

\end{abstract}

\section{INTRODUCTION}

In the real world, most tasks are long-horizon in nature, such as tidying rooms, conducting experiments, or assembling equipment, which requires robots to have the ability to stably execute extended sequences of actions. In recent years, LLM-empowered robots have demonstrated exceptional capabilities enabliing them to perform numerous tasks\cite{mu2024embodiedgpt}\cite{chen2024vlmimic}\cite{wake2024gpt}\cite{chen2025fmimic}. However, when faced with long-horizon tasks, especially those involving complex scenarios, LLM planners often produce suboptimal results.

In order to apply LLMs to long-horizon tasks, previous works have been proposed step-by-step method such as FLTRNN\cite{sermanet2024robovqa}, LLM-State\cite{sharan2024plan}. These methods effectively improve the faithfulness and success rate of LLMs in complex long-horizon tasks by integrating task decomposition and memory management into the planning process. However, these methods predominantly rely on LLMs for direct reasoning, and are thus constrained by the inherent reasoning limitations of these models, resulting in performance that deteriorates significantly as task complexity increases.

To address the aforementioned performance degradation in long-horizon planning scenarios, recent research, inspired by the Tree of Thought (ToT) framework\cite{yao2023tree}, has applied diverse sampling and plan selection strategies to embodied planning, including Tree-Planner\cite{yao2024tree} and RAP\cite{hao2023reasoning}. These approaches construct tree-like structures by sampling various plans and selecting optimal paths at each decision node. Although such methods have demonstrated improved success rates and enhanced flexibility in long-horizon tasks, they nonetheless fundamentally attempt to generate comprehensive plans in a single instance. Despite the optimization process of selecting the optimal outcomes from multiple generations produced by LLMs, this limitation persists because LLMs are still applied directly to long sequential reasoning tasks at each inference step.

Despite the progress that has been made, reliably converting long-horizon tasks into a sequence of actions remains a formidable challenge. In long-horizon tasks, performance failures primarily stem from two factors: (1) \textbf{Contextual Gap}: Long task sequences inherently introduce redundant information into the input context, a phenomenon that becomes significantly more pronounced in long-horizon tasks. With ratio of redundant context increased in the context, it negatively impacts reasoning capabilities. A more pronounced contextual gap diminishes the planner's capacity to comprehend critical task-relevant information, thereby impairing its inferential performance. (2) \textbf{Logical gap}: In task planning contexts, extended sequences typically correspond with increasingly abstract linguistic instructions. Consequently, the inferential process from abstract instruction directives to concrete executable steps becomes more challenging. Short tasks, such as 'picking up an object,' can be easily associated with specific actions like 'grasping' whereas more complex tasks such as 'cleaning the table' are relatively difficult to connect directly to primitive actions. A more pronounced logical gap correlates with diminished inferential success rates.

Based on the preceding analysis, mitigating logical and contextual gaps is essential for enhancing robotic performance in long-horizon task planning. 
To achieve this objective, we present \underline{\textbf{S}}ubgoal \underline{\textbf{T}}ree \underline{\textbf{E}}mbodied \underline{\textbf{P}}lanner (\textbf{STEP}). Unlike tree structures employed in the ToT method, which independently constructs temporal action sequences along each branch, the tree constructed by STEP features subgoals at each level derived from the preceding layer, systematically building the tree through logical decomposition, with leaf nodes serving as action sequences. This approach thereby connects instructions and actions through the subgoal tree, mitigating the logical gap between them. Additionally, due to the coarse-to-fine decomposition characteristic of the tree structure, using parent nodes rather than the overall task as input context for the planner reduces redundant context. This enables the planner to focus specifically on the decomposition process at each step, thereby reducing the contextual gap.

To construct the subgoal tree structure, we have developed a closed-loop framework comprising: 
(1) A \textbf{subgoal decomposition model} that recursively breaks down complex tasks into precise, discrete subtasks organized in a hierarchical tree structure. 
(2) A \textbf{leaf node termination model} that evaluates whether each decomposed node satisfies task requirements and can be directly mapped to a specific action, thereby determining when to terminate the subgoal tree spanning.

This approach effectively reduces the logical and contextual gaps in the planning process by enabling the planner to focus on the decomposition or termination of current subgoals. 

Our main contributions are threefold:

(I) We present \textbf{STEP}, a framework in long-horizon reasoning for embodied tasks using foundation LLMs, which mitigates the contextual and logical gap. This approach enables LLMs to perform long-chain reasoning more efficiently for embodied tasks.

(II) We propose a pair of closed-loop \textbf{subgoal decomposition model} and \textbf{leaf node termination model} for the spanning and termination of the subgoal tree, facilitating the dynamic construction of the subgoal tree.

(III) Extensive experiments demonstrate SOTA in both VirtualHome and real-robot systems across four different kinds of manipulation tasks, with up to 34\% success rates(WAH-NL) and 25\%(real robot) outperforming previous SOTA methods.

\begin{figure*}[h]
\centering
\includegraphics[width=\linewidth]{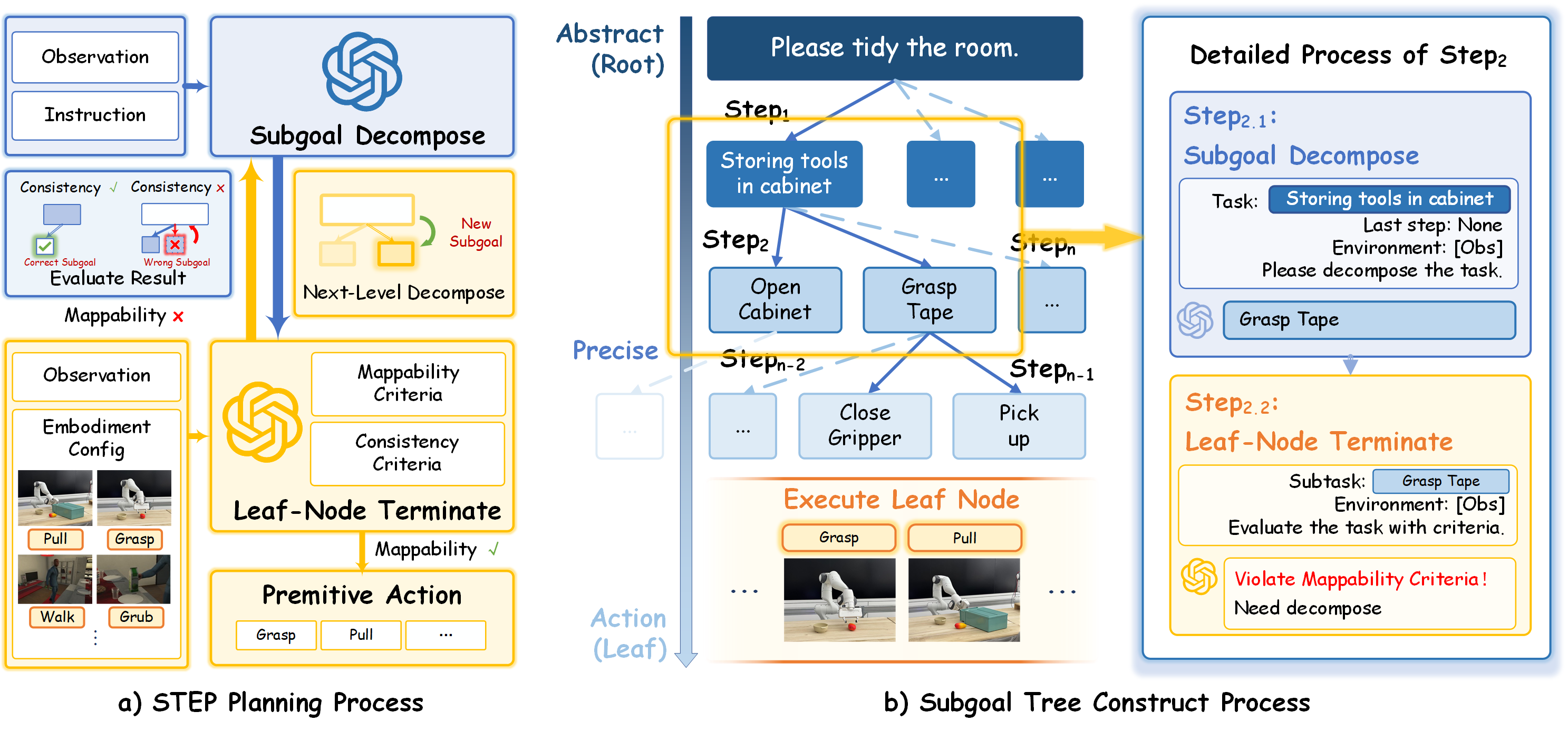}
\caption{\textbf{Framework of STEP.} \textbf{a)} STEP constructs a subgoal tree through a closed-loop process of subgoal decomposition and leaf-node termination. In the subgoal decomposition model, the LLM breaks down complex tasks into subtasks. Each subgoal is then evaluated against mappability and consistency criteria to determine whether it should be further decomposed, re-planned, or executed directly. \textbf{b)} STEP progressively refines subgoals into more precise actions and determines the next executable action for each newly generated subgoal (left) through iterative subgoal decomposition and leaf node termination (right). This recursive process generates a complete subgoal tree, as illustrated using the progression from step 2 as an example.}
\label{fig:1}
\end{figure*}

\section{RELATED WORKS}
\subsection{LLMs as Embodied Planner}
LLMs, having been trained on vast multimodal datasets, possess extensive high-level prior knowledge about real-world environments. With the development of imitation learning, robots are now able to perform short tasks more effectively\cite{chen2025unify}\cite{black2024pi_0}\cite{chen2025graphMimic}. Previous research has demonstrated the utility of LLMs as embodied planners, enabling long-horizon tasks through the invocation of relevant skill libraries and the combination of skill functions\cite{vemprala2024chatgpt}\cite{yao2022react}. This approach effectively leverages the models' prior knowledge to translate human natural language instructions into executable sequences of robotic actions. However, this method frequently encounters challenges due to mismatches between planned sequences and real-world constraints or affordances, resulting in reduced task success rates.

To address these challenges, researchers have employed reinforcement learning to evaluate the plan\cite{ahn2022can}, while others have adopted the method of self-feedback based on LLMs, allowing them to reflect on the correctness of their plan\cite{shinn2024reflexion}\cite{lin2024swiftsage}. Additionally, external environmental feedback can be introduced to assess the scenario after task execution\cite{huang2022inner}\cite{cui2023human}. The reinforcement learning approach can effectively determine the accuracy of planning based on a value function, but it requires a substantial amount of data to obtain a robust network for action assessment. The external feedback approach, by directly incorporating environmental feedback, enables LLMs to intuitively understand execution outcomes, thereby reducing internal reasoning processes\cite{bhat2024grounding}. However, its real-time effectiveness is limited, as evaluations can only be conducted post-execution, restricting its range of applications.

\subsection{Long-chain Inference with LLMs}
LLMs have demonstrated proficiency in tasks that involve planning and reasoning. In this regard, researchers have begun to investigate the potential of LLMs for long chain planning. Studies relying solely on foundational LLM have shown inadequate performance in task completion. Using GPT-4 turbo directly as a planner in Travel Planner benchmark, results in a poor final pass rate\cite{achiam2023gpt}. To address this issue, some researchers employ re-ranking techniques\cite{wang2022self} and iterative correction\cite{shinn2024reflexion}\cite{madaan2024self} to mitigate the inherent dilemmas faced by LLMs. Nonetheless, these two methodologies have not fundamentally resolved the intrinsic difficulty for LLMs functioning as planners.

Meanwhile, some researchers are using the coding capabilities of LLMs by transforming natural language tasks into abstract code generation\cite{chaffin2021ppl}\cite{gu2022don}. These methodologies enable LLMs to effectively undertake long planning tasks by transforming them into reasoning tasks that LLMs are more adept at handling. However, tree-based construction approaches require LLMs to engage in common sense acquisition tasks, whereas code-based or other logical expression transformation methods limit planning applications in open environments due to their predefined primitives and functions. Consequently, both approaches face difficulties adapting to complex environments and tasks, which restricts the range of planner applications.

\section{METHOD}

In this research, we propose STEP as a framework that enables embodied agents to decompose high-level tasks into sequences of primitive actions for real-world long-horizon tasks. In the following sections, we first analyze the planning gaps inherent in embodied task planning and introduce the \textbf{subgoal tree} architecture (Sec. \ref{sec:IIIA}). Subsequently, we present the \textbf{subgoal decomposition model}, which systematically decomposes high-level tasks into executable subtasks with reduced logical and contextual gaps (Sec. \ref{sec:IIIB}). We then detail our approach for evaluating subgoal tree leaf nodes through the development of \textbf{leaf node termination model} (Sec. \ref{sec:IIIC}). Through iterative refinement of this task decomposition framework, we aim to transform abstract, complex instructions into concrete, executable action sequences, thereby enabling robotic systems to effectively adapt to and accomplish diverse tasks in real-world environments.

\subsection{Subgoal Tree} 
\label{sec:IIIA}
\textbf{Preliminaries.} In typical LLM planners, the planner is defined as a Partially Observable Markov Decision Process (POMDP), represented as $\langle\mathcal{S},\mathcal{O},\mathcal{A},\mathcal{T}\rangle$, where $\mathcal{S}$ denotes a set of states, $\mathcal{O}$ a set of observations, $\mathcal{A}$ a set of actions, and $\mathcal{T}$ a transition model. In practical inference, the process of generating action sequences is described as:
\begin{align}\label{eq:eq1}\pi_\phi(a_t|g,h_t,o_t).\end{align}

This process utilizes the LLMs' reasoning capabilities to generate the current state's action $a_t$ based on the goal \textit{g}, historical actions $h_{t}=\left\{a_{0}, \cdots, a_{t-1}\right\}$, and the current observation state $o$. 

However, this methodology faces a fundamental theoretical challenge: the substantial \textbf{contextual} and \textbf{logical gaps} that exist between high-level instructions and the corresponding low-level executable actions.	

\textbf{Contextual Gap.} From a contextual perspective, foundation models exhibit suboptimal coherence when processing and generating long context due to their attention-based architecture. This limitation stems from attention reduction that occurs with prolonged contextual input\cite{shi2023large}, compromising internal consistency. Similarly, in embodied contexts, extended task sequences lead to greater redundant information input, as LLMs must process both human instructions and the complete history of executed actions to generate the current action. This results in substantial irrelevant context being inputted, causing attention reduction, which leads to lower planning success rates for long-term tasks.

\textbf{Logical Gap.} From a logical perspective, next-token prediction methodologies encounter significant limitations as foundation models are not explicitly trained to transform abstract knowledge into detailed actions during their learning phase\cite{bachmann2024pitfalls}. This constraint impedes their ability to effectively ground abstract natural language instructions in specific executable steps, resulting in suboptimal performance when planning complex embodied tasks. Consequently, the substantial gap between high-level instructions and requisite detailed actions contributes to diminished success rates in generating comprehensive plans for long-horizon tasks.

\textbf{Subgoal Tree.} 
To address the aforementioned issues, we propose the subgoal tree, a hierarchical structure that systematically progresses from coarse to fine granularity, effectively reducing both logical and contextual gaps during the planning process. In the subgoal tree, human instructions serve as the root node, which is progressively decomposed into multiple child nodes during the planning process. Each child node represents a subgoal of its parent node, whereby the effect of each parent node can be achieved through the cumulative effects of all its child nodes. This decomposition process continues recursively until each leaf node corresponds to a permitive action. At this point, the sequence of leaf nodes represents the precise actions to be executed, with the cumulative effects of all leaf nodes aligning with the intended effect of the root node, thereby ensuring the reliability of the decomposition.

By reducing the logical gap between hierarchical levels, this methodology enhances the rationality and coherence of each inferential process. Through the implementation of single-level reasoning workflows, the planner leverages both the contextual outputs from its parent node and the sequential reasoning history within the current level, thereby capitalizing on the parent task's inherent encompasses of higher-order mission objectives, result in lower logical gaps.

Consequently, the hierarchical decomposition framework ensures that historical actions from disparate branches of the subgoal tree are systematically excluded from the input context of the current planning process. This architectural constraint effectively transforms the task required decomposition into a refined, contextually-relevant representation of the essential input information. By eliminating redundant contextual inputs, this approach enables the LLM planner to concentrate with greater precision on the specific decomposition task, thereby yielding demonstrably superior planning outcomes.

\subsection{Subgoal Decomposition Model}
\label{sec:IIIB}

We formulate the decompose model as a POMDP. In a POMDP setting, the observation $o_t$ represents a subset of the underlying state $s_t$. Instead of  Eq.\eqref{eq:eq1}
where $g$ represent the goal in natural language, $h_t$ as the history of actions and $o_t$ as the observation, the process of subgoal decomposition can be described as: 
\begin{align}
\label{eq:eq2}
\pi_{tree}({\Phi^{n}(\{b_i\}\vert_{i=0}^n)} \vert {\Phi^{n-1}(\{b_i\}\vert_{i=0}^{n-1})},{\Phi^{n}(\{b_{n}\}\vert_{i=0}^{n-1},b_{i-1})},o_t).
\end{align}
where $\Phi$ denote the goal decomposition operator, $b_i$ represent as the $b_i$-th branch of the $ (i-1)$-th level subgoal ($b_i$ = 0, 1, 2\dots), ${\Phi^{n}(\{b_i\}\vert_{i=0}^n)}$ denotes the $b_n$-th subgoal of branch \{$b_0$,$b_1$,...,$b_{n-1}$\}. The generated subgoal is determined by the parent node, left-side node, and current environmental observation. Consequently, we can deduce that:
\begin{align}
{\Phi^{n}(\{b_i\}\vert_{i=0}^n)}=\Phi({\Phi^{n-1}(\{b_i\}\vert_{i=0}^{n-1}}),b_n).
\end{align}which shows the fundamental subgoal decompose process. 

\begin{figure}[!t]
\includegraphics[width=\linewidth]{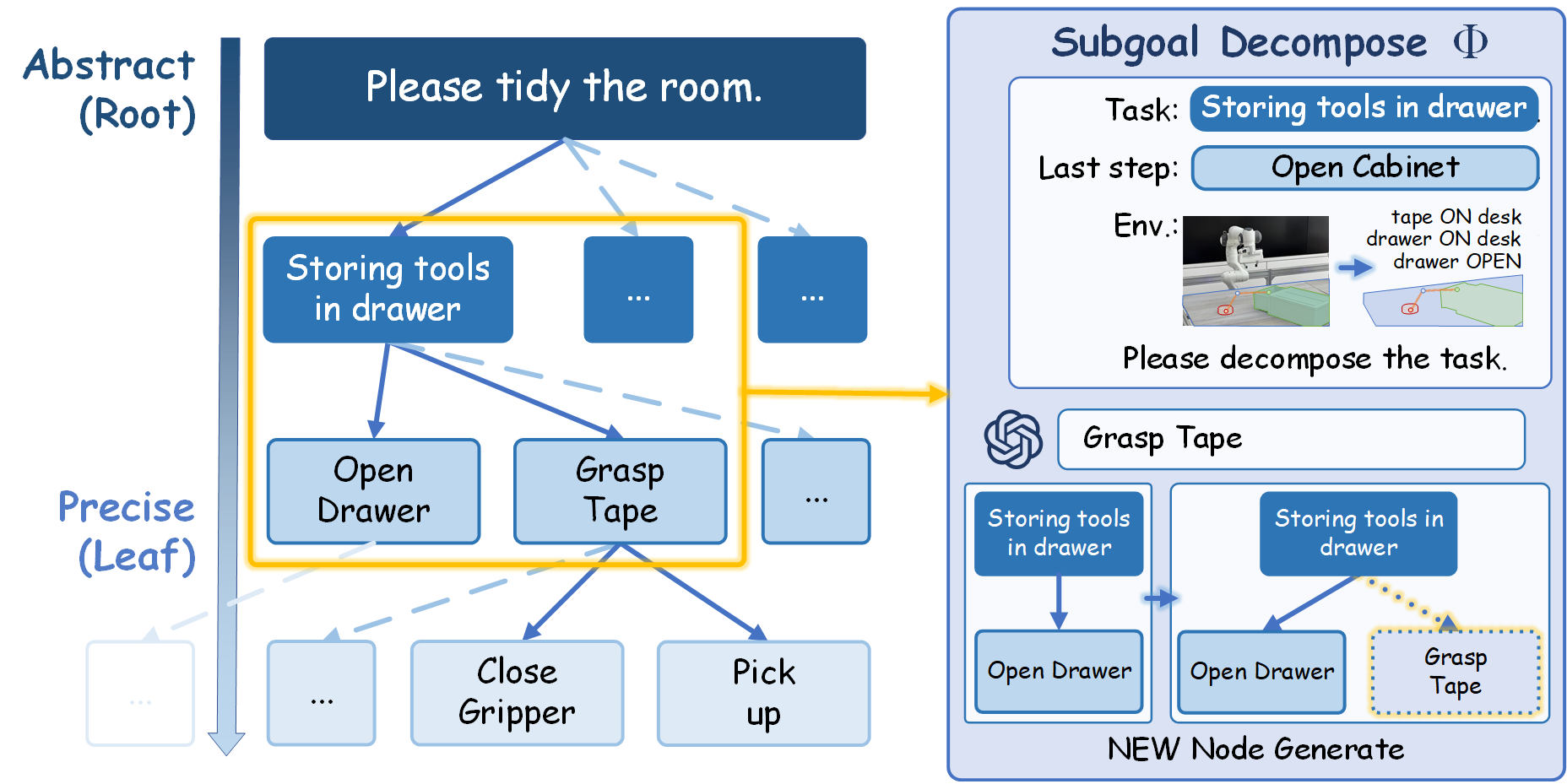}
\caption{\textbf{Framework of subgoal decompose model.} Taking decomposing “Storing tools in drawer” as an example, the subgoal decomposition model can directly infer the next subgoal to be generated from the parent node and the previous step, i.e., grasp tape.}\label{fig:3}
\centering
\end{figure}

The specific process is illustrated in the Fig.\eqref{fig:3}. In each step, the planner obtains the current observation $o_t$, the results of planning at a higher level ${\Phi^{n-1}(\{b_i\}\vert_{i=0}^{n-1})}$, and the previous step at the same level ${\Phi^{n}(\{b_{n}\}\vert_{i=0}^{n-1},b_{i-1})}$. 

\subsection{Leaf Node Termination Model}
\label{sec:IIIC}
To facilitate the termination of the subgoal tree decomposition process, we have devised an embodiment-related leaf node termination model. This model is designed to stop further refinement when the level of detail achieved through decomposition precisely enables the resultant components to be mapped onto a primitive action.

Within the leaf node termination model, the current subtask generated by the subgoal tree, along with the higher-level task requirements, the previous subtasks at the same level, and the embodiment configuration are entered into the LLM. The LLM then evaluates every subtask output from the subgoal tree using the aforementioned assessment process.

Consequently, we posit that an embodiment-related leaf node termination model must fulfill two essential functions, which include:

1. \textbf{Mappability Criteria}. Assess whether the task can be mapped into an explicit primitive action. The criteria are shown as:
\begin{equation}
\begin{aligned}
\forall	a_i \ne a_j, f(\Phi_T(i)) \to a_i, f(\Phi_T(j)) \to a_j, \\
\Phi_T(i) \ne \Phi_T(j).
\end{aligned}
\end{equation}

2. \textbf{Consistency Criteria}. Determine whether the current subgoal satisfies the embodiment's affordance, aligns with the requirements of the previous subgoal, and is executable in the current environment.

\textbf{Affordance}. 
\textcolor{myColor}{Assessment of subtask compatibility with the embodiment's affordances}, denote $\mathcal{A}(\mathcal{E})$ as the affordance of the embodiment $\mathcal{E}$, $\mathcal{A}$ denote the leaf node of subgoal tree as actions $\mathcal{A} = <a_0 \land a_1 \cdots \land a_n>$ , $\Phi_T $ as a leaf node. The criteria are shown as:

\begin{equation}
\begin{aligned}
\forall	a_i \in \mathcal{A}, f(\Phi_T(i)) \to a_t, \\
a_i \in \mathcal{A}(\mathcal{E}).
\end{aligned}
\end{equation}

For example, in certain scenarios, LLMs may overlook the capability constraints of the current embodiment, resulting in the planning of unfeasible tasks. A case in point would be a single arm being instructed to perform additional tasks with its end effector while it is already holding an object. Consequently, a well-designed leaf node termination model should incorporate functionality to evaluate the feasibility of proposed plans, encompassing both the current embodiment's ability to execute the plan and the plan's viability within the existing environment.

\textbf{Task Congruence}. \textcolor{myColor}{Evaluation of subtask executability within the current environmental constraints, determining whether implementation can proceed without further decomposition}. The criteria are shown as:

\begin{equation}
\begin{aligned}
\forall	\Phi^{n}(\{b_i\}\vert_{i=0}^n) \ \in T,\\
{\Phi^{n}(\{b_i\}\vert_{i=0}^n)}=\Phi({\Phi^{n-1}(\{b_i\}\vert_{i=0}^{n-1}}),b_n).
\end{aligned}
\end{equation}

\begin{figure}[!t]
\includegraphics[width=\linewidth]{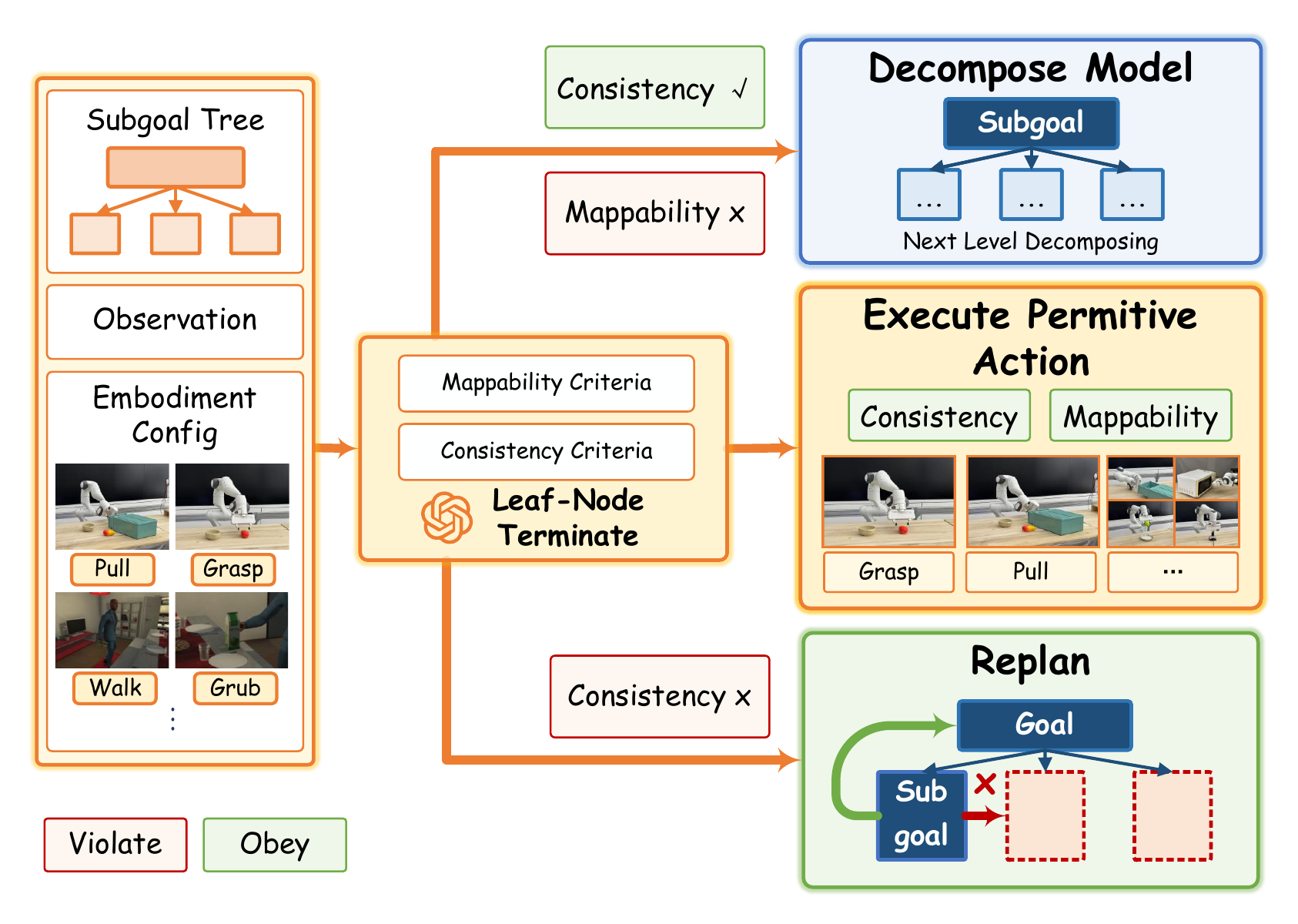}
\caption{\textbf{Framework of leaf node termination model.} Mappability criteria and consistency criteria are used to determine whether the current node should execute a primitive action, execute the next decomposition, or replan.}
\label{fig:4}
\centering
\end{figure}

\textbf{Environment}. \textcolor{myColor}{Evaluation of subtask executability within the current environmental constraints, determining whether implementation can proceed without further decomposition}, denote $\mathcal{A}(\mathcal{S})$ as the constraint of state, $T_t$, denote the leaf node of $T$ as the trajectory $T_t=<a_0 \land a_1 \cdots \land a_n>$. The criteria are shown as:

\begin{equation}
\begin{aligned}
\forall	a_i \in \mathcal{A}, f(\Phi_T(i)) \to a_t, \\
a_i \in \mathcal{A}(\mathcal{S}).
\end{aligned}
\end{equation}
The logical structure of the criterion is illustrated in Fig.\eqref{fig:4}.
\begin{itemize}
    \item If a task satisfies both of the criteria, it is mapped to a macro action and a success status is returned.
    \item If a task does not meet the mappability criterion but satisfies the consistency criteria, instructions for further refinement are returned.
    \item If a task fail to meet the consistency criteria, a planning error is returned, and the previous round of subgoal planning is re-initiated.
\end{itemize}
In summary, the subgoal tree construction process is described in Algorithm\eqref{alg:AOA}. 
\begin{algorithm}[!h]
    \caption{Algorithm of \textit{subgoal tree} construction}
    \label{alg:AOA}
    \renewcommand{\algorithmicrequire}{\textbf{Input:}}
    \renewcommand{\algorithmicensure}{\textbf{Output:}}
    
    \begin{algorithmic}[1]
            \REQUIRE Instruction $goal$, Observation $obs$ 
        \ENSURE SubgoalTree $T$  
        
        \STATE subgoal = $goal$
        \STATE Tree.addNode(subgoal)
        \COMMENT {\textit{Initialize subgoal tree}}
        \WHILE{evaluate(Tree.allLeafNode) is not mappable}
            \STATE subgoal = decompose(subgoal, $obs$)
            
            \COMMENT {\textit{Decompose current subgoal}}
            \IF {evaluate(subgoal) is mappable}
                \STATE execute(subgoal.action)
                
                \COMMENT {\textit{Execute after successfully decomposed}}
                \IF {allNeighbourComplete(subgoal)}
                    \STATE subgoal = subgoal.parent.next
                    
                    \COMMENT {\textit{Return to parent subgoal after complete}}
                \ELSE
                    \STATE subgoal = subgoal.next
                    
                    \COMMENT {\textit{Move on to next subgoal}}
                    
                \ENDIF
            \ELSE
                \STATE subgoal = subgoal.parent
                
                \COMMENT {\textit{Replan parent node when error occurs}}
            \ENDIF
        \ENDWHILE
    \end{algorithmic}
\end{algorithm}

\section{EXPERIMENT}
\subsection{Evaluation on VirtualHome Simulator}
\label{sec:IVA}
\begin{figure*}[h]
\centering
\includegraphics[width=\linewidth]{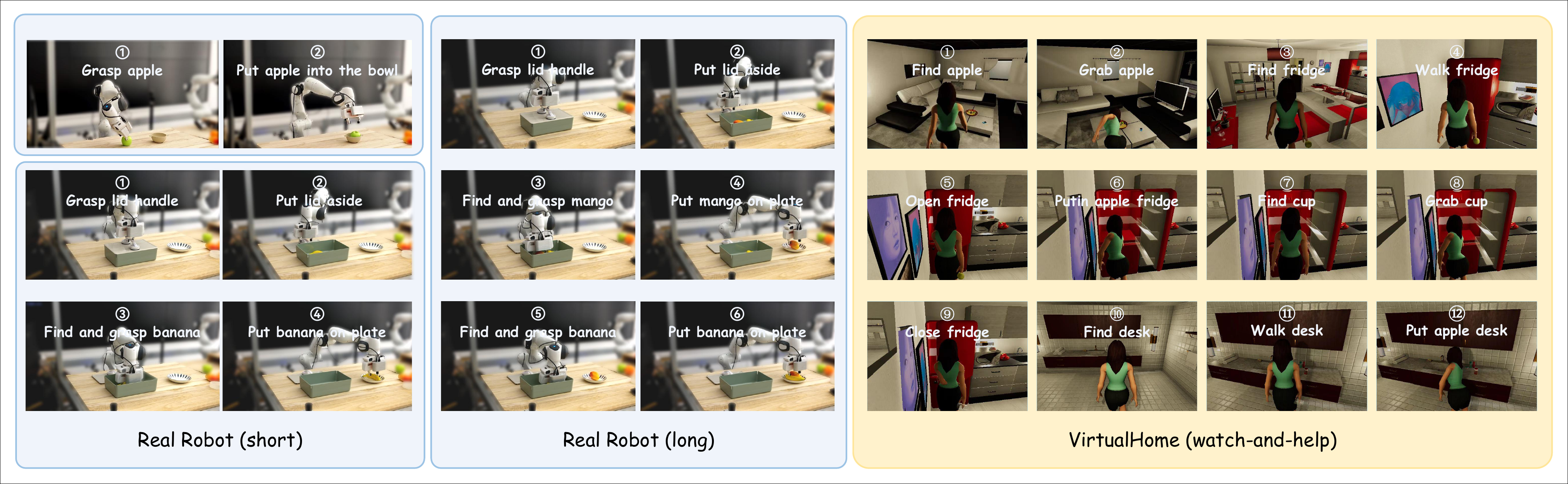}
\caption{\textbf{Planning results}. The experiment was tested on the Virtual Home simulator and real robots, demonstrating the planning performance of STEP in complex environments and long tasks.}
\label{fig:5}
\end{figure*}
\textbf{Environment.}
We conducted the experiments in the VirtualHome environment \cite{puig2018virtualhome}, a simulation platform designed for complex household tasks. In the experiment, we used the WAH-NL benchmark \cite{choi2024lota}, which is designed to convert natural language instructions into a sequence of actions in real-life scenarios. This benchmark presents challenges for robotic task planning due to the lengthy nature of each task step, the richness of objects within the scene, and the logical dependencies between objects (for example, get in-cabinet objects must be retrieved by opening cabinets first).

\textbf{Dataset.}
We conducted tests on the annotated dataset of 100 entries in the WAH-NL benchmark, which consists of 350 labeled robot datasets. Each data set includes at least one natural language instruction, an initial scene state, and at least one subgoal completion criterion. If all subgoal criteria are met, the overall task goal is considered accomplished. Following the evaluation, we obtained the subgoal success rate(success rate of individual subgoals) and success rate(success rate of overall tasks).

\textbf{Baselines.}
As an experiment, we used the WAH-NL benchmark to compare the Saycan\cite{ahn2022can}, LoTa-Bench\cite{choi2024lota}, and ProgPrompt\cite{singh2023progprompt} methods, and analyzed the failure modes of the three methods separately. Based on the EAI\cite{li2024embodied}, we classified them as \textit{Grammer error}, \textit{Goal Satisfaction Error}(including Missing State, Missing Relation, Missing Goal Action) and \textit{Trajectory Runtime Error} (including Wrong Order, Affordance Error, Additional/Missing Step).

\textbf{Main Results.}
According to the results in Table\eqref{table1}, several advantages of STEP can be obtained:

\begin{table}[h]
\centering
\caption{Performance of different methods on WAH-NL benchmark.}
\label{table1}
\resizebox{\linewidth}{!}{
\fontsize{8}{10}\selectfont
\begin{tabular}{m{2.60cm}<{\centering} *{2}{m{2.50cm}<{\centering}}}
\Xhline{1pt}
\textbf{Method}  & \textbf{SR.$\uparrow$} & \textbf{SSR.$\uparrow$}\\ 
\Xhline{0.5pt}
SayCan\cite{ahn2022can}  & 1\% & 2.07\% \\
ProgPrompt\cite{singh2023progprompt}  & 3\% & 18.69\% \\
LoTa-Bench\cite{choi2024lota}  & 6\% & 36.79\% \\
\Xhline{0.5pt}
\rowcolor{gray!10} 
\textbf{STEP(Ours)}  & \textbf{40\%} & \textbf{62.02\%} \\
\Xhline{1pt}
\end{tabular} 
}
\label{MRFsum}
\end{table}

(i) STEP performed better than the baseline systems in WAH-NL, with an increase 34\% increased in success rates and 25.23\% increased in subgoal success rates. The empirical performance of LoTa-Bench and SayCan demonstrates similar outcomes to those reported in \cite{choi2024lota}.

(ii) The experiment showed that in the WAH-NL benchmark, the improvement ratio in the success rate was much higher than that in the subgoal success rate, indicating that STEP has a more significant effect on tasks with longer time series, suggesting that the task decomposition method based on the subgoal tree can better cope with long-horizon tasks.

\textbf{Result Analysis.}
Through the analysis of various baseline failure results in Table\eqref{table2}.
\begin{table*}[t]
    \centering
    \caption{Error Type Analysis of Different Methods on WAH-NL Benchmark.}
    \label{table2}
    \resizebox{\linewidth}{!}{
\fontsize{8}{10}\selectfont
\begin{tabular}{m{2.00cm}<{\centering} |*{3}{m{2.00cm}<{\centering}} |*{3}{m{2.00cm}<{\centering}}}
        \Xhline{1pt}
        \raisebox{-3mm}[0pt][0pt]{\textbf{Method}} & 
        \multicolumn{3}{c|}{\textbf{Goal Satisfaction Error}} & 
        \multicolumn{3}{c}{\textbf{Trajectory Runtime Error}} \\ 
        \cline{2-7}
        & \textbf{Missing State} & \textbf{Missing Relation} & \textbf{Missing Goal Action} & \textbf{Wrong Order} & \textbf{Additional/ Missing Step} & \textbf{Affordance Error}  \\ 
        \Xhline{0.5pt}
        Saycan\cite{ahn2022can} &   5\% & 7\% & 9\% & 9\% & 32\% & 15\%  \\ 
        ProgPrompt\cite{singh2023progprompt} &   2\% & 6\% & 10\% & 11\% & 36\% & 12\%  \\ 
        LoTa-Bench\cite{choi2024lota} &   3\% & 3\% & 9\% & 28\% & 36\% & 13\%  \\ 
        \Xhline{0.5pt}
        \rowcolor{gray!10} 
        \textbf{STEP (Ours)}  & \textbf{1\%} & \textbf{2\%} & \textbf{2\%} & \textbf{2\%} & \textbf{27\%} & \textbf{3\%} \\ 
        \Xhline{1pt}
    \end{tabular}
    }
    \label{ErrorTypeAnalysis}
\end{table*}
STEP outperforms all other baselines in terms of \textit{Goal Satisfaction Error} and \textit{Trajectory Runtime Error} across all levels. STEP's primary error type is categorized as "Additional/Missing Steps," which stems from limitations in the LLM-based leaf-node termination model. This limitation arises because the leaf-node termination model relies on LLMs, and despite their capabilities, foundation models still demonstrate relatively limited environmental comprehension, leading to potential misjudgments in certain scenarios. This constitutes the primary bottleneck for further performance improvements in the STEP framework.

Additionally, there is no significant difference in the rate of \textit{Grammar Error} between STEP(15\%) and the baseline methods(16.3\% avg.). This can be attributed to the fact that both methods utilize the same backbone, with the primary source of \textit{Grammar Error} stemming from the hallucination of LLMs. As a result, different schemes exhibit similar rates of \textit{Grammar Error}. 

\subsection{Evaluation on Real Robot}
For the underlying planner, we employed RoboScript API\cite{chen2024roboscript} and deployed it on the actual robot. We categorized the tasks into four levels. 
\begin{itemize}
    \item The "short-simple" category comprises fewer than 5 steps with all objects being visible.
    \item The "short-complex" category comprises fewer than 5 steps but includes hidden objects (e.g., items in drawers).
    \item The "long-simple" category comprises 5 to 8 steps with all objects being visible.
    \item The "long-complex" category comprises 5 to 8 steps and includes hidden objects.

\end{itemize}
Each group consists of 2 tasks, and each task is tested 5 times, using the baseline from Sec.\ref{sec:IVA} for tests.

\textbf{Environment Setup.}
In real robot experiments, we selected the Franka Emika Panda and utilized OpenAI's GPT-4o as the LLM backbone. For the underlying planner, we employed the API from RoboScript\cite{chen2024roboscript}, which handled the low-level motion planning and control, and successfully deployed this integrated system to the actual robot.

\textbf{Main Results.}
As shown in Table\eqref{table3}, in real robot experiments, we observed that in simple tasks, the performance of STEP was similar to that of different baselines. In more complex tasks, as the task length increased, STEP demonstrated superior performance.
As the complexity of the tasks increased (requiring the identification of unseen objects), the performance advantage of STEP became more pronounced.
Even when using the same LLM backbone, STEP performed better in tasks that required stronger reasoning capabilities.

\begin{table}[!h]
    \centering
    \caption{Success Rate of different methods on Real Robots.}
    \label{table3}
    \resizebox{\linewidth}{!}{
\fontsize{8}{10}\selectfont
\begin{tabular}{m{2.00cm}<{\centering} *{4}{m{1.30cm}<{\centering}}}
    \Xhline{1pt}   
    \textbf{Method} & \textbf{short-simple} & \textbf{short-complex} & \textbf{long-simple} & \textbf{long-complex}\\ 
    \Xhline{0.5pt}
    Saycan\cite{ahn2022can} & 8/10 & 6/10 & 4/10 & 1/10 \\
    ProgPrompt\cite{singh2023progprompt} & 9/10 & 8/10 & 3/10 & 1/10\\
    RoboScript\cite{chen2024roboscript} & 9/10 & 8/10 & 5/10 & 1/10\\
    LoTa-Bench\cite{choi2024lota} & 10/10& 8/10& 6/10& 1/10\\
    \Xhline{0.5pt}
    \rowcolor{gray!10} 
    \textbf{STEP(Ours)}& \textbf{10/10}& \textbf{10/10}&\textbf{ 9/10}& \textbf{6/10} \\
    \Xhline{1pt}
    \end{tabular}
    }
    \label{MRFsum}
\end{table}
\subsection{Ablation Study}
To evaluate the contribution of different components of STEP to planning success rates, we designed the following ablation experiments:

(1) \textbf{w/o tree structure:} In this condition, we modify the input information for each task by replacing the parent node with both the original parent node and all leaf nodes executed previously. This modification eliminates the condensing effect of the tree structure on the current task while maintaining the similar logical gap. The resulting redundant information allows us to specifically investigate how the tree structure's reduction of the contextual gap contributes to improved success rates.

(2) \textbf{w/o subgoal tree:} In this condition, we modify each task's input information by replacing the parent node with the root node and all previously executed leaf nodes. This approach eliminates the subgoal tree structure, removing both the task condensation effect and the reasoning difficulty reduction achieved through multi-step reasoning in the STEP method. By maintaining a contextual gap similar to the "-w/o tree" group, this condition allows us to investigate the specific impact of subgoal task condensation on task planning effectiveness.

Evaluate the aforementioned approach within Sec.\ref{sec:IVA}.

\textbf{Results Analysis.} The evaluation results as depicted in Table\eqref{table4} demonstrate that the success rates of STEP with ablated subgoal and tree structures are diminished compared to the full STEP implementation. 

\begin{figure}[h]
\centering
\includegraphics[width=0.8\linewidth]{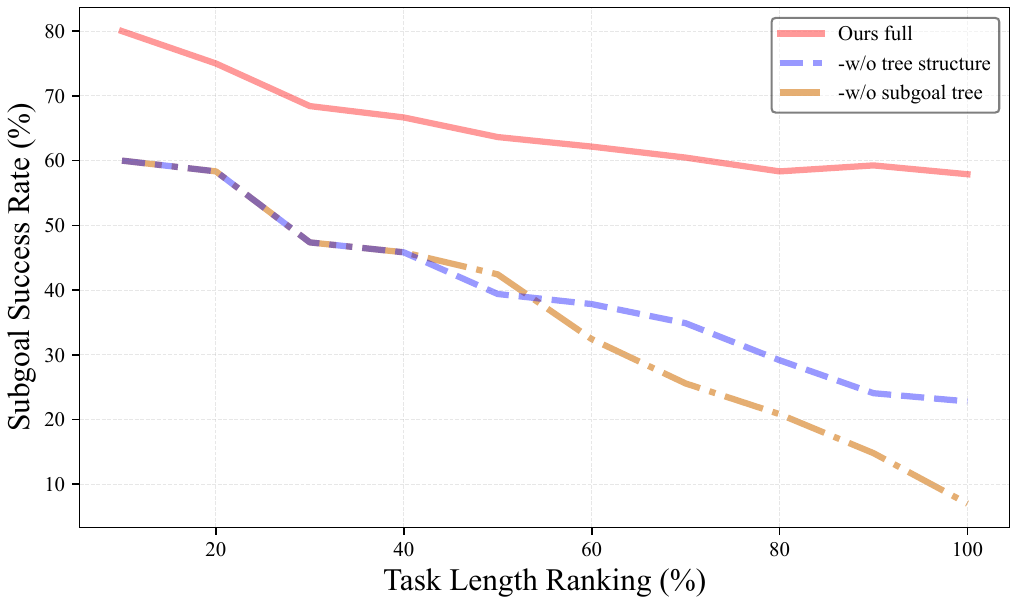}
\caption{\textbf{Task Length and Subgoal Success Rate Variation under Different Baselines.} In the WAH benchmark, we group tasks by length into 10\% intervals, from short to long, and calculate the subgoal success rate for each task within the intervals.
}
\label{fig:6}
\end{figure}

As illustrated in Fig.\eqref{fig:6}, reveals that all methods experience a decline in success rates as task length increases. 
\begin{table}[h]
\centering
\caption{Performance of different methods on WAH-NL benchmark.}
\label{table4}
\resizebox{\linewidth}{!}{
\fontsize{8}{10}\selectfont
\begin{tabular}{m{2.60cm}<{\centering} *{2}{m{2.50cm}<{\centering}}}
\Xhline{1.2pt}
\textbf{Method} & \textbf{SR.} & \textbf{SSR.}\\ 
\Xhline{0.7pt}
\textbf{Ours full}  & \textbf{40\%} & \textbf{62.02\%} \\
-w/o tree structure  & 8\% & 34.12\% \\
-w/o subgoal tree  & 9\% & 27.30\% \\
\Xhline{1.2pt}
\end{tabular} 
}
\label{MRFsum}
\end{table}

As observed, the "-w/o tree structure" configuration demonstrates inferior performance compared to the "full STEP" approach, while the "-w/o subgoal tree" configuration performs even worse than the "-w/o tree structure" ones. This evidence substantiates that the subgoal tree structure's capacity to reduce both logical and contextual gaps collectively contributes to STEP's enhanced effectiveness.

Notably, the method with the ablated subgoal tree exhibits a higher rate of decline, attributable to the subgoal tree's principle of reducing the larger logical gap between abstract instructions and actions in more complex tasks. Consequently, as tasks become longer, the impact of the logical gap on performance becomes increasingly pronounced, substantiating that the subgoal tree effectively diminishes the logical gap in the planning process. 

\section{CONCLUSIONS}
In this paper, we present STEP, a general approach that enables LLMs to adapt to long-horizon embodied planning tasks. By constructing a subgoal tree, STEP reduces the contextual and logical gaps in embodied planning tasks, enabling LLMs to function effectively as robot planners for complex embodied tasks. STEP first decomposes the task using a subgoal decomposition model to construct a subgoal tree layer by layer. It then employs a leaf-node termination model to determine when to terminate the spanning of each branch. Upon branch completion, the corresponding action is executed. Experiments conducted in both simulators and real robots demonstrate that STEP significantly outperforms baseline approaches in embodied tasks, particularly in long-horizon scenarios and complex environments. In summary, STEP expands the applicability of LLMs to robotic embodied tasks, enabling robots to better execute long-horizon human instructions in open-world environments.

\addtolength{\textheight}{-12cm}   

 \section*{ACKNOWLEDGMENT}
We would like to express our sincere appreciation to Mr. Yijun Yuan from Beijing Institute of Technology for his help during the experimental process. 

\bibliography{ref}

\end{document}